\documentclass[conference]{IEEEtran}
\IEEEoverridecommandlockouts

\usepackage{cite}
\usepackage{amsmath,amssymb,amsfonts}
\usepackage{algorithmic}
\usepackage{graphicx}
\usepackage{textcomp}
\usepackage{xcolor}
\usepackage{color}
\usepackage{booktabs} 
\usepackage{url}      
\usepackage{hyperref} 
\usepackage[table]{xcolor} 
\definecolor{rowA}{RGB}{230,244,255} 
\definecolor{rowB}{RGB}{255,239,230} 
\definecolor{rowC}{RGB}{234,247,239} 
\definecolor{rowD}{RGB}{243,233,253} 
\definecolor{rowE}{RGB}{255,249,230} 
\definecolor{rowF}{RGB}{255,234,239} 
\definecolor{rowG}{RGB}{236,243,252} 
\definecolor{rowH}{RGB}{232,248,245} 
\definecolor{rowI}{RGB}{245,236,250} 
\definecolor{rowJ}{RGB}{250,241,232} 
\usepackage{afterpage}
\begin{document}

\title{DUAL-VAD: Dual Benchmarks and Anomaly-Focused Sampling for Video Anomaly Detection}

\author{
\IEEEauthorblockN{Seoik Jung}
\IEEEauthorblockA{\textit{AI/Data Team} \\
\textit{PIA-SPACE Inc.}\\
si.jung@pia.space}
\and
\IEEEauthorblockN{Taekyung Song}
\IEEEauthorblockA{\textit{AI/Data Team} \\
\textit{PIA-SPACE Inc.}\\
tg.song@pia.space}
\and
\IEEEauthorblockN{Joshua Jordan Daniel}
\IEEEauthorblockA{\textit{Dept. of Artificial Intelligence and Robotics} \\
\textit{Sejong University}\\
jordandj2001@gmail.com}

\and
\IEEEauthorblockN{JinYoung Lee\textsuperscript{*}}
\IEEEauthorblockA{\textit{Dept. of Artificial Intelligence and Robotics    } \\
\textit{Sejong University}\\
jinyounglee@sejong.ac.kr}
\and
\IEEEauthorblockN{SungJun Lee\textsuperscript{*}}
\IEEEauthorblockA{\textit{AI/Data Team} \\
\textit{PIA-SPACE Inc.}\\
sj.lee@pia.space}
}


\maketitle
\begingroup\def\thefootnote{*}\footnotetext{Corresponding authors.}\endgroup

\begin{abstract}
Video Anomaly Detection (VAD) is critical for surveillance and public safety.
However, existing benchmarks are limited to either frame-level or video-level tasks, restricting a holistic view of model generalization.
This work first introduces a softmax-based frame allocation strategy that prioritizes anomaly-dense segments while maintaining full-video coverage, enabling balanced sampling across temporal scales.
Building on this process, we construct two complementary benchmarks.
The image-based benchmark evaluates frame-level reasoning with representative frames, while the video-based benchmark extends to temporally localized segments and incorporates an abnormality scoring task.
Experiments on UCF-Crime demonstrate improvements at both the frame and video levels, and ablation studies confirm clear advantages of anomaly-focused sampling over uniform and random baselines.
\end{abstract}


\section{Introduction}

Video Anomaly Detection (VAD) is a long-standing and fundamental research challenge with critical importance in surveillance and public safety. Existing approaches can be broadly divided into two paradigms.

The first is \textbf{frame-level evaluation}, where each frame is assessed individually for abnormality. Recent works have shown that such evaluation can even be conducted \textit{without explicit training}, by treating anomaly detection as an image-based inference problem \cite{vera,holmesvad}.

The second is \textbf{video-level classification}, where an entire video is treated as a unit and judged normal or abnormal. This typically involves dividing the video into temporal segments and aggregating anomaly scores across them \cite{stead,ucfcrime}.

While both paradigms have advanced the field, there has been no unified framework that simultaneously supports them and systematically emphasizes anomaly-dense regions. Moreover, existing benchmarks remain limited in scope, often failing to include high-level reasoning cases that are crucial for real-world security scenarios \cite{ucvl,sharegpt4video}.

\noindent\textbf{Why sampling matters.}
In fully/weakly supervised VAD, training or evaluating on all frames would be ideal for preserving anomaly evidence, but the time/compute cost is prohibitive for long videos.
Practical alternatives such as uniform sampling, random sampling, or keyframe-based heuristics tend to under-sample the critical moments where anomalies occur, resulting in diluted supervision and unstable evaluation.
Therefore, designing sampling strategies that reflect anomaly likelihood is essential for effective training and evaluation.

To address these gaps, we propose a framework that first applies a Video Segmentation Module to divide videos into coherent units, and then assigns anomaly likelihoods using an Anomaly Scorer. Building upon this process, we make the following three contributions:

\begin{enumerate}
    \item \textbf{Anomaly-Focused Sampling:} We propose a softmax-based frame allocation strategy, where 
    frames are allocated across segments in proportion to their anomaly likelihoods. 
    This strategy emphasizes anomaly-rich regions while maintaining coverage of the entire video.
    \item \textbf{Dual Benchmarks for VAD:} Using the proposed sampling strategy, we construct both an image-based benchmark and a video-based benchmark from the same segmentation–scoring pipeline. The image benchmark enables frame-level reasoning, while the video benchmark supports video-level classification with abnormality scoring. Both benchmarks include not only basic recognition tasks but also challenging reasoning-oriented questions, extending evaluation toward more challenging, reasoning-oriented scenarios.
    \item \textbf{Empirical Validation:} Experiments on UCF-Crime demonstrate state-of-the-art performance at the video level and strong, competitive performance at the frame level (VERA protocol). We also release our benchmarks and models as open-source resources to encourage reproducibility and future research.
\end{enumerate}

Through this unified process, our work establishes both a principled evaluation protocol and an effective training strategy for anomaly detection across modalities and temporal scales.

\section{Related Work}

\subsection{Video Anomaly Detection (VAD)}
Video anomaly detection has been widely studied and is commonly grouped into \textbf{frame-level detection} and \textbf{video-level classification}. Frame-level approaches evaluate anomalies per frame and can be formulated as image inference without additional training using vision–language models \cite{vera,holmesvad,vador,lavad}. Video-level approaches instead divide videos into temporal segments and aggregate anomaly scores for classifying a whole video \cite{ucfcrime,rtfm,mgfn2021,s3r2022,claws2021,mist2020}.
To reduce annotation cost, weakly supervised and MIL-style methods have been explored \cite{rtfm,wsal2019,claws2021,mist2020,pel2022,urdmu2022,bnwvad2023}. However, most prior work still adheres to a single paradigm, and benchmarks that \emph{jointly} support both levels remain scarce.

\subsection{Vision–Language Models for Video Understanding}
Recent multimodal large language models (MLLMs) have opened new opportunities for video understanding by combining visual encoders with language reasoning \cite{sharegpt4video,clip,llava,imagebind}. While these studies report strong results on tasks such as video QA and captioning, the anomaly-detection setting—characterized by rare-event recognition and the need for explainable decisions—has been much less explored. Our work explicitly aligns VLM-based explainability with both frame- and video-level anomaly evaluation \cite{vera,holmesvad,vador}.

\subsection{Benchmarks and Datasets}
Large-scale resources such as UCF-Crime contain long, untrimmed surveillance videos with temporally localized abnormal events and are widely adopted for both frame- and video-level evaluation \cite{ucfcrime}. Beyond anomaly datasets, unified video–language benchmarks emphasize instruction-following and higher-order reasoning \cite{ucvl,sharegpt4video}. Shot-level segmentation datasets and models further demonstrate the effectiveness of partitioning videos into coherent units for efficient understanding \cite{autoshot}. Nevertheless, most existing datasets emphasize \emph{either} frame- or video-level analysis. In contrast, we construct \emph{both} an image-based (frame-level) and a video-based benchmark from a single segmentation–scoring pipeline, to enable consistent evaluation across temporal scales and difficulty levels.

\subsection{Sampling Strategies for Video Understanding}
Sampling is a critical component in training efficient video models. Uniform or random sampling is simple but often under-represents anomaly-dense regions; alternatives like keyframe extraction or shot-based segmentation improve efficiency but remain insensitive to anomaly distribution \cite{autoshot}. We address this by introducing an \textbf{anomaly-focused sampling} strategy that allocates frames in proportion to segment-level anomaly scores, emphasizing abnormal segments without sacrificing global coverage.

\section{Method}

\subsection{Overview}
We formalize a unified pipeline for constructing dual benchmarks and training models with anomaly-focused sampling.  
Given a video represented as a sequence of frames 
\[
V = \{v_1, v_2, \dots, v_T\}, 
\]
our method first divides $V$ into shot segments, assigns anomaly scores, and then applies a softmax-based frame allocation strategy. The allocated frames are used to construct both image- and video-based benchmarks, which in turn support model training and evaluation.

\begin{figure}[t]
    \centering
    \includegraphics[width=\linewidth]{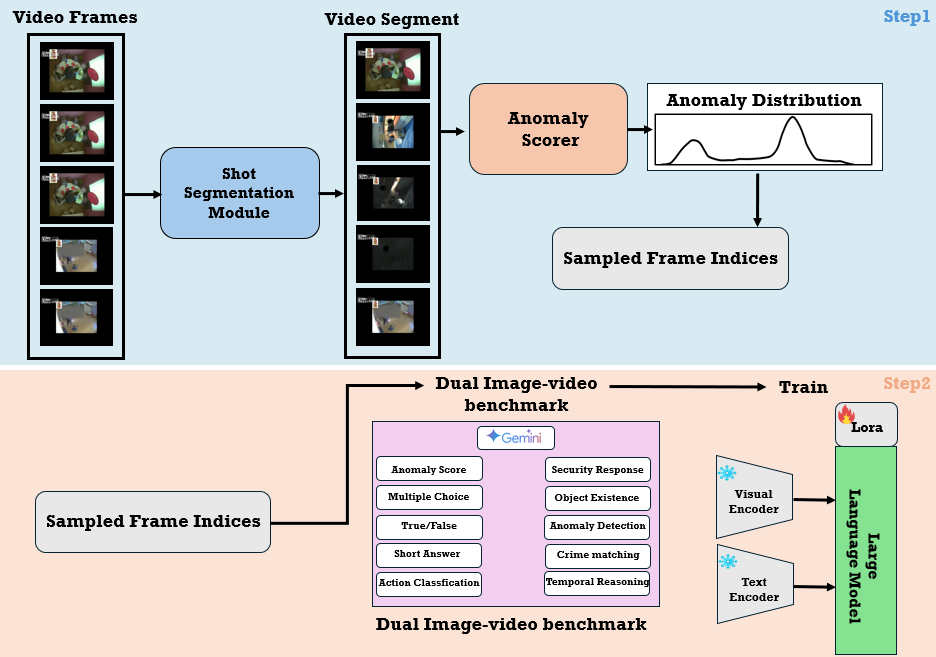}
    \caption{Overall framework. A video $V$ is segmented into $\{S_i\}_{i=1}^M$ by the \textbf{Shot Segmentation Module} and scored by the \textbf{Anomaly Scorer}, producing scores $\{s_i\}$. Frames are allocated using anomaly-focused sampling and then used to build dual benchmarks (image-level and video-level). Both benchmarks generate structured QA pairs for training and evaluation.}
    \label{fig:overview}
\end{figure}

\begin{table*}[t]
\centering
\caption{Overview of the 10 question types included in our dual benchmarks.}
\label{tab:question_types}
\renewcommand{\arraystretch}{1.18} 
\resizebox{\textwidth}{!}{
\begin{tabular}{p{3.0cm}p{4.2cm}p{4.2cm}p{5.0cm}}
\toprule
\textbf{Question Type} & \textbf{Description} & \textbf{Key Evaluation Focus} & \textbf{Example Question} \\
\midrule
\rowcolor{rowA}
Multiple Choice & 4-option question on person/object/action/situation & Visual identification, situational reasoning & ``What is this person doing?'' \\
\rowcolor{rowB}
True/False (Binary Judgment) & Yes/No answer to a declarative statement & Factual reasoning, time/location recognition & ``Was this scene captured in a restricted area?'' \\
\rowcolor{rowC}
Short Answer & Brief descriptive response about judgment or security action & Decision-making, inference & ``What should the security officer do in this situation?'' \\
\rowcolor{rowD}
Action Classification & Classify or select the action of a person & Pose recognition, interaction analysis & ``What is the person in the red shirt doing?'' \\
\rowcolor{rowE}
Object Existence & Ask if a specific object is present & Object detection ability & ``Is there a firearm visible in the frame?'' \\
\rowcolor{rowF}
Spatial Reasoning & Determine relative position of objects/persons & Spatial awareness & ``Is the person located on the left side of the frame?'' \\
\rowcolor{rowG}
Anomaly Detection & Decide whether the scene is abnormal and why & Contextual reasoning, norm awareness & ``Does this scene appear to be abnormal?'' \\
\rowcolor{rowH}
Crime Type Matching & Match abnormal scenes to crime categories & Scene-to-concept classification & ``If abnormal, what type of crime does this indicate?'' \\
\rowcolor{rowI}
Temporal Reasoning & Infer temporal context from visual clues & Time-related interpretation & ``Was this scene captured during nighttime hours?'' \\
\rowcolor{rowJ}
Security Response & Decide most appropriate response action & Decision-making, tactical planning & ``What is the most appropriate security response in this situation?'' \\
\bottomrule
\end{tabular}
}
\end{table*}

\subsection{Image-based Benchmark Construction}
For frame-level reasoning, we design an \textbf{image-based benchmark}.  

\begin{enumerate}
    \item \textbf{Shot segmentation:} The video $V$ is partitioned into $M$ segments using a pre-trained shot segmentation model~\cite{autoshot}.
    \[
    V = \{S_1, S_2, \dots, S_M\}, \quad S_i = \{v_{a_i}, \dots, v_{b_i}\}.
    \]
    \item \textbf{Anomaly scoring:} We use the STEAD~\cite{stead} model as our Anomaly Scorer to assign a score $s_i \in [0,1]$ to each segment $S_i$.
    \item \textbf{Frame extraction:} For the specific purpose of generating QA pairs for the benchmark, we employ a simple heuristic to extract three frames from abnormal segments ($s_i \geq 0.5$) and one from normal segments:
    \[
    n_i = 
    \begin{cases}
    3, & s_i \geq 0.5, \\
    1, & s_i < 0.5.
    \end{cases}
    \]
    \item \textbf{Question generation:} Sampled frames are passed to teacher models (Gemini Pro, Gemini Flash) to generate structured QA pairs in JSON schema.  
    Recent studies emphasize that high-quality captions significantly enhance multimodal reasoning~\cite{sharegpt4v}, motivating our use of strong teacher models.
\end{enumerate}

\subsection{Video-based Benchmark Construction}
For video-level reasoning, we extend the same pipeline to construct a \textbf{video benchmark}.  

\begin{enumerate}
    \item \textbf{Shot segmentation \& scoring:} As above, we obtain $\{S_i\}$ and scores $\{s_i\}$.
    \item \textbf{Labeling:} Segments are labeled as abnormal if $s_i \geq 0.5$, otherwise normal.
    \item \textbf{Question generation:} Teacher models generate QA pairs incorporating anomaly context.  
    This design aligns with recent multimodal large language models for video understanding~\cite{gpt4video}, which highlight instruction-followed reasoning and safety-aware evaluation.
    \item \textbf{Expanded tasks:} In addition to recognition and reasoning questions, we include \textbf{Abnormality Scoring} tasks, requiring a continuous anomaly score $\hat{y} \in [0,1]$ for the full video.
\end{enumerate}

In total, our constructed benchmarks contain 6,130 frame-level QA pairs and 2,840 video-level QA pairs, 
resulting in 8,970 samples across anomaly types.

\subsection{Anomaly-Focused Frame Allocation}
To sample $N$ frames from $M$ segments, we allocate frames according to segment-level anomaly scores $\{s_i\}$.  

\textbf{Step 1. Softmax weighting}
\[
p_i = \frac{\exp(s_i / \tau)}{\sum_{j=1}^M \exp(s_j / \tau)}
\]

\textbf{Step 2. Raw allocation}
\[
\hat{n}_i = N \cdot p_i
\]

\textbf{Step 3. Constraint enforcement}
\[
n_i = \min\Big(\max(\lfloor \hat{n}_i \rfloor,\, 1),\, \mathrm{max}_i\Big)
\]

\textbf{Step 4. Residual adjustment}  
If $\sum_i n_i \neq N$, the residual is redistributed based on descending order of $(\hat{n}_i - n_i)$.

This ensures every segment contributes at least one frame while anomaly-rich segments are prioritized.

\subsection{Training and Evaluation Setup}

\noindent\textbf{Model Backbone.}  
We adopt InternVL3-2B as the vision–language backbone, following the design of recent multimodal large language models~\cite{internvl,sharegpt4video,gpt4video}. 
The visual encoder $\phi_v$ is frozen and inherits the ViT structure from CLIP~\cite{clip}. 
Given an input video $V \in \mathbb{R}^{T \times H \times W \times C}$ with frames $\{\mathcal{V}_i\}_{i=1}^T$, each frame is mapped into visual tokens:
\[
V_i = \{ v_i^{cls}, v_i^1, \dots, v_i^{N_p} \} = \phi_v(\mathcal{V}_i),
\]
where $v_i^{cls}$ is the class token and $N_p$ is the number of patch tokens.  
The text encoder $\phi_t$ tokenizes and embeds the input question $Q$ into $X_q = \phi_t(Q)$.

\noindent\textbf{Anomaly-Focused Sampling.}  
Unlike prior works that adopt ATS-based frame selection~\cite{holmesvad,zhou2023dual}, 
we apply our anomaly-focused frame allocation strategy (Sec.~III-D) to select a subset of informative frames $\mathcal{G} \subseteq \{1,\dots,T\}$.  
The selected tokens are projected into the language space by a projector $\phi_p$:  
\[
X_{ins} = \text{cat}\,[\, \phi_p(\text{cat}[V_i]),\, X_q \,], \qquad i \in \mathcal{G}.
\]

\noindent\textbf{Training Objective.}  
The concatenated embeddings $X_{ins}$ are fed into the LLM to predict the answer sequence $X_a$.  
The likelihood of generating the full sequence is:
\[
p(X_a \mid X_{ins}) = \prod_{i=1}^{L} p_\theta(x_i \mid X_{ins,<i}, X_{a,<i}),
\]
where $\theta$ are trainable parameters and $L$ is the answer length.  
We optimize the standard autoregressive cross-entropy loss, which is widely used in multimodal LLM training~\cite{llava,sharegpt4v}:
\[
\mathcal{L}_{QA} = - \sum_{i=1}^{L} \log p_\theta(x_i \mid X_{ins,<i}, X_{a,<i}).
\]

\noindent\textbf{Fine-tuning Strategy.}  
To adapt the backbone efficiently without disrupting pretrained capabilities, we employ LoRA-based parameter-efficient fine-tuning~\cite{lora}.  
Only the projector $\phi_p$ and LoRA adapters in the LLM are updated, while the visual encoder $\phi_v$ remains frozen.

\noindent\textbf{Evaluation Protocol.}  
We evaluate under two canonical paradigms:
\begin{itemize}
    \item \textbf{Frame-level:} Report AUC on sampled frames following the VERA protocol~\cite{vera}.
    \item \textbf{Video-level:} Aggregate predicted abnormality scores to classify full videos.
\end{itemize}

\section{Experiments}

\subsection{Experimental Setup}
All experiments are conducted on the UCF-Crime \cite{ucfcrime} dataset, a widely adopted benchmark for video anomaly detection. 
Following the protocol established in the UCA-Crime \cite{uca} and later extended by Holmes-VAU \cite{holmesvau}, 
we adopt UCF-Crime as the \emph{base dataset} for constructing our training data and evaluation benchmarks.

\begin{table}[htbp]
\centering
\caption{Video-level anomaly classification results on UCF-Crime. The 'DUAL\_VAD (Ours)' entry represents the base model fine-tuned on both the image and video benchmarks.}
\label{tab:stead_results}
\begin{tabular}{lc}
\toprule
\textbf{Method} & \textbf{AUC-ROC (\%)} \\
\midrule
Sultani et al. (C3D) \cite{ucfcrime} & 75.41 \\
Sultani et al. (I3D) \cite{ucfcrime} & 77.92 \\
MIST (I3D) \cite{mist2020} & 82.30 \\
CLAWS (C3D) \cite{claws2021} & 83.03 \\
RTFM (VideoSwin) \cite{rtfm} & 83.31 \\
RTFM (I3D) \cite{rtfm} & 84.03 \\
WSAL (I3D) \cite{wsal2019} & 85.38 \\
S3R (I3D) \cite{s3r2022} & 85.99 \\
MGFN (VideoSwin) \cite{mgfn2021} & 86.67 \\
PEL (I3D) \cite{pel2022} & 86.76 \\
UR\text{-}DMU (I3D) \cite{urdmu2022} & 86.97 \\
MGFN (I3D) \cite{mgfn2021} & 86.98 \\
BN\text{-}WVAD (I3D) \cite{bnwvad2023} & 87.24 \\
Anomaly Scorer (Fast, X3D) \cite{stead} & 88.87 \\
\textbf{Anomaly Scorer (Base, X3D)} \cite{stead} & \textbf{91.34} \\
\midrule
\textbf{DUAL\_VAD (Ours)} & \textbf{92.33} \\
\bottomrule
\end{tabular}
\end{table}

To comprehensively evaluate anomaly detection, we follow two canonical paradigms:
\begin{itemize}
    \item \textbf{Frame-level evaluation:} Following the vision-language evaluation protocol (as in VERA \cite{vera}), where frame-level AUC is measured without additional model training.
    \item \textbf{Video-level evaluation:} Standard video-level classification, where AUC is computed across full videos using anomaly scores.
\end{itemize}

For preprocessing, videos are divided into segments using a video segmentation module, and each segment is assigned an anomaly score by an anomaly scorer. 
Unless stated otherwise, frame sampling follows our anomaly-focused allocation strategy (Sec. III-D).

\begin{table}[htbp]
\centering
\caption{Frame-level anomaly detection results on UCF-Crime. The 'DUAL\_VAD (Ours)' entry represents the model fine-tuned solely on the image benchmark.}
\label{tab:vera_results}
\begin{tabular}{lc}
\toprule
\textbf{Method} & \textbf{AUC (\%)} \\
\midrule
\multicolumn{2}{l}{\textit{Non-explainable VAD Methods}} \\
Wu et al.~\cite{wu2020} & 82.44 \\
OVVAD~\cite{ovvad2023} & 86.40 \\
S3R~\cite{s3r2022} & 85.99 \\
RTFM~\cite{rtfm} & 84.30 \\
MSL~\cite{msl2020} & 85.62 \\
MGFN~\cite{mgfn2021} & 86.98 \\
SSRL~\cite{ssrl2021} & 87.43 \\
CLIP-TSA~\cite{cliptsa2022} & 87.58 \\
Sultani et al.~\cite{ucfcrime} & 77.92 \\
GCL~\cite{gcl2020} & 79.84 \\
GCN~\cite{gcn2019} & 82.12 \\
MIST~\cite{mist2020} & 82.30 \\
CLAWS~\cite{claws2021} & 83.03 \\
DYANNET~\cite{dyannet2019} & 84.50 \\
Tur et al.~\cite{tur2020} & 66.85 \\
GODS~\cite{gods2021} & 70.46 \\
\midrule
\multicolumn{2}{l}{\textit{Explainable VAD Methods}} \\
LAVAD~\cite{lavad} & 80.28 \\
Holmes-VAD~\cite{holmesvad} & 84.61 \\
VADor~\cite{vador} & 85.90 \\
ZS CLIP~\cite{clip} & 53.16 \\
ZS IMAGEBIND-I~\cite{imagebind} & 53.65 \\
ZS IMAGEBIND-V~\cite{imagebind} & 55.78 \\
LLAVA-1.5~\cite{llava} & 72.84 \\
Frame-level VLM Eval (VERA)~\cite{vera} & 86.55 \\
\midrule
\textbf{DUAL\_VAD (Ours)} & \textbf{85.77} \\
\bottomrule
\end{tabular}
\end{table}

\subsection{Main Results}
The main experimental results are summarized in Table \ref{tab:stead_results} for video-level evaluation and Table \ref{tab:vera_results} for frame-level evaluation.
Our method, \textbf{DUAL\_VAD}, achieves a new state-of-the-art at the video level, surpassing prior CNN- and transformer-based methods. At the frame level, our approach also demonstrates competitive performance, validating the effectiveness of anomaly-focused sampling and dual benchmark training across both paradigms.

\subsection{Benchmark Contribution Analysis}
To validate the contribution of our dual benchmarks, we compare models trained with different benchmark configurations: the image-based benchmark only, the video-based benchmark only, and the combined benchmark. 
Results are shown in Table \ref{tab:benchmark_analysis}.

\begin{table}[htbp]
\centering
\caption{Performance comparison of benchmark configurations on UCF-Crime (ROC-AUC).}
\label{tab:benchmark_analysis}
\resizebox{\columnwidth}{!}{%
\begin{tabular}{lcc}
\toprule
\textbf{Training Data} & \textbf{Video-level} & \textbf{Frame-level} \\
\midrule
Base (InternVL3-2B) & 0.812 & 0.777 \\
Base + Image Benchmark & 0.919 & \textbf{0.857} \\
Base + Video Benchmark & 0.908 & 0.833 \\
Base + Image + Video & \textbf{0.923} & 0.823 \\
\bottomrule
\end{tabular}
}
\end{table}

We observe that the \textbf{image-based benchmark} yields the largest individual improvement over the base model, boosting video-level AUC from 0.812 to 0.919 (+0.107) and frame-level AUC from 0.777 to 0.857 (+0.080). This suggests that frame-level QA pairs provide relatively strong and less noisy supervision. The \textbf{video-based benchmark} also improves performance (0.908 / 0.833), though to a slightly lesser extent (+0.096 / +0.056), likely due to its smaller scale and the added difficulty of temporal reasoning.

When both benchmarks are combined, the model achieves the highest video-level performance (0.923), indicating a clear complementary effect across temporal and spatial reasoning. However, the frame-level score (0.823), while higher than the base, does not surpass the image-only (0.857) or video-only (0.833) settings. This suggests that while integration strengthens video-level generalization, image-based supervision remains more directly effective for frame-level tasks.

\begin{table}[htbp]
\caption{Ablation study on sampling strategies (Video-level AUC).}
\label{tab:sampling_ablation}
\centering
\begin{tabular}{lc}
\toprule
\textbf{Sampling Strategy} & \textbf{AUC-ROC (\%)} \\
\midrule
Uniform Sampling & 87.24 \\
Random Sampling & 87.88 \\
\textbf{Anomaly-focused Sampling (Ours)} & \textbf{87.92} \\
\bottomrule
\end{tabular}
\end{table}

\subsection{Ablation on Sampling Strategies}
To isolate the impact of each sampling strategy, we conducted this ablation by fine-tuning the base model on only the video benchmark derived from UCF-Crime \cite{holmesvau, uca}. All other conditions were held constant. This controlled setup explains why the AUC scores in Table \ref{tab:sampling_ablation} differ from our main results, which leveraged both image and video benchmarks for training.

The results of the three settings are reported below:
\begin{itemize}
    \item \textbf{Uniform sampling:} Frames are evenly sampled across the video.
    \item \textbf{Random sampling:} Frames are selected randomly from each segment.
    \item \textbf{Proposed anomaly-focused sampling:} Frames are allocated based on anomaly scores.
\end{itemize}

Our anomaly-focused strategy shows a consistent advantage over both baselines, demonstrating that emphasizing anomaly-dense segments improves downstream detection performance.

\subsection{Open-source Contribution}
To facilitate reproducibility and community advancement, we publicly release both the constructed \textbf{image- and video-based benchmarks} as well as the trained models. We believe these resources will serve as valuable foundations for future research in video anomaly detection.

\section{Conclusion}
We presented \textbf{DUAL-VAD}, a unified framework that (i) allocates frames with a \emph{softmax-based anomaly-focused} strategy to emphasize anomaly-dense segments while preserving global coverage, and (ii) constructs \emph{dual benchmarks}—an image-based benchmark for frame-level reasoning and a video-based benchmark for temporal reasoning and abnormality scoring—from a single segmentation–scoring pipeline.

On UCF-Crime, DUAL-VAD achieves a new \textbf{state of the art at the video level} with \textbf{AUC-ROC 92.33\%}, surpassing the strongest prior baseline (91.34\%). At the frame level (VERA protocol), our model attains 85.77\% AUC, competitive with recent explainable VAD systems, validating that anomaly-focused selection and the proposed benchmarks improve both recognition and reasoning.

Ablations further clarify where the gains arise. Training with the \emph{image-based benchmark} yields the largest boost for frame-level performance (AUC $0.857$), while combining \emph{image + video} data delivers the best video-level result (AUC $0.923$) with strong frame-level accuracy. Our sampling study shows a consistent edge for \emph{anomaly-focused allocation} over uniform and random baselines (87.92\% vs. 87.24\% / 87.88\% AUC-ROC), supporting the value of aligning sampling density to anomaly likelihood.

However, some limitations still remain: naive joint training can introduce mild supervision conflicts that slightly dampen frame-level scores compared to image-only training. Future work will explore curriculum or multi-task weighting to better balance the two benchmarks, extend evaluation beyond UCF-Crime, and refine segment scoring to further close the frame-level gap.

We will release the benchmarks and code to facilitate reproducible research and to encourage broader adoption of unified, reasoning-oriented evaluation for video anomaly detection.

\section*{Acknowledgments}
This research was supported by the Institute of Information
\& Communications Technology Planning \& Evaluation (IITP)
grant funded by the Korea government (MSIT) (RS-2022-00156345).


\end{document}